\documentclass{article} 
\usepackage{colm2024_conference}

\usepackage{microtype}
\usepackage{hyperref}
\hypersetup{
    colorlinks=true,
    linkcolor=blue,
    citecolor=blue,
    urlcolor=blue,
}
\usepackage{url}
\usepackage{wrapfig,lipsum,booktabs} 
\usepackage{enumitem}
\usepackage{amsmath, bm}
\usepackage{xcolor, colortbl}
\usepackage{pifont}
\usepackage{multirow,multicol}
\usepackage{mwe}
\usepackage{arydshln,booktabs, caption}
\usepackage{color,soul}
\usepackage{graphicx}
\usepackage[utf8]{inputenc}
\usepackage{CJKutf8} 

\newcommand{\hlc}[2][yellow]{{%
    \colorlet{foo}{#1}%
    \sethlcolor{foo}\hl{#2}}%
}

\title{Towards Chapter-to-Chapter Context-Aware Literary \\ Translation via Large Language Models}

\author{Linghao Jin \& Li An \& Xuezhe Ma 
\\
Information Sciences Institute\\
University of Southern California\\
\texttt{\{linghaoj,lan72605,xuezhema\}@usc.edu}
}

\colmfinalcopy 
\begin{document}

\maketitle

\begin{abstract}
Discourse phenomena in existing document-level translation datasets are sparse, which has been a fundamental obstacle in the development of context-aware machine translation models.
Moreover, most existing document-level corpora and context-aware machine translation methods rely on an unrealistic assumption on sentence-level alignments. 
To mitigate these issues, we first curate a novel dataset of Chinese-English literature, which consists of 160 books with intricate discourse structures.
Then, we propose a more pragmatic and challenging setting for context-aware translation, termed chapter-to-chapter (\textsc{Ch2Ch}) translation, and investigate the performance of commonly-used machine translation models under this setting.
Furthermore, we introduce a potential approach of finetuning large language models (LLMs) within the domain of \textsc{Ch2Ch} literary translation, yielding impressive improvements over baselines. 
Through our comprehensive analysis, we unveil that literary translation under the \textsc{Ch2Ch} setting is challenging in nature, with respect to both model learning methods and translation decoding algorithms.
\end{abstract}

\section{Introduction}
Despite the efforts on developing context-aware machine learning systems to meaningfully exploit inter-sentential information, recent work has investigated the fundamental obstacles in existing document-level translation datasets and context-aware machine translation models~\citep{jin-etal-2023-challenges}.
First, existing datasets lack the necessary contextual information and/or discourse phenomena for meaningful document-level translation~\citep{lupo-etal-2022-divide}. 
Second, existing predominant context-aware translation methods assume that  sentence-level alignments are available during training, which does not accurately represent real-world translation scenarios~\citep{thai-etal-2022-exploring,jin-etal-2023-challenges}. 

To remedy the issues, recent work has pivoted to literary translation and proposed a more realistic paragraph-to-paragraph setting, given that literary texts typically contain complex discourse structures that
mandate a document-level frame of reference.
\citet{thai-etal-2022-exploring} released \textsc{Par}3, a paragraph-level translation dataset sourced from recently-published 118 novels in 19 languages (about 6 novels per language on average).
\citet{jin-etal-2023-challenges} curated \textsc{Para2Para}, a small-scale dataset consisting of 10,545 parallel paragraphs across six novels.
However, these datasets are either in small scale or the reference translations are automatically generated from machine translation systems (e.g. Google Translate~\citep{Wu2016GooglesNM} and fine-tuned GPT-3~\citep{brown2020language}).
In addition, there still exist some serious limitations in the paragraph-to-paragraph translation setting, including limited contextual information and equivocal paragraph splits in literary texts.

Large language models (LLMs) with decoder-only Transformer architectures have demonstrated outstanding performance as sentence-level translation systems~\citep{vilar2023prompting, jiao2023chatgpt, kocmi2023large, zhang2023bayling, yang2023bigtranslate}.
In the aspect of context-aware translation, recent studies have employed decoder-only LLMs to translate entire paragraphs using few-shot in-context learning methods, yielding impressive translation quality~\citep{karpinska-iyyer-2023-large}. 
However, how to finetune LLMs to process context-aware translation for literary texts in a more realistic and challenging scenario remains under-explored. 

In this paper, we propose a more pragmatic and challenging setting for context-aware translation, named \emph{chapter-to-chapter} (\textsc{Ch2Ch}), associated with a carefully curated dataset of Chinese-English literature.
The dataset consists of 160 literary books, together with professional translations in Chinese. 
Then we investigate the performance of commonly-used machine translation models under the proposed setting and dataset.
In addition, we investigate the efficacy of applying LLMs in context-aware chapter-to-chapter literary translation and highlight several key challenges that impede the progress. 
Our main contributions are outlined as follows:

\begin{itemize}[leftmargin=*]
    \item We propose a more realistic setting for literary translation: chapter-to-chapter(\textsc{Ch2Ch}) translation, wherein a document is translated at the granularity of chapters. To support it, we release a chapter-aligned Chinese-English dataset (JAM), comprising 5,373 parallel chapters extracted from 160 novels, to catalyze future research endeavors.
    \vspace{-1mm}
    \item Through comprehensive analysis, we unveil the challenges in chapter-level translation, including long-context model training and decoding strategies.
    \vspace{-1mm}
    \item With empirical experiments, we evaluate the performance of recent trending LLMs on the JAM dataset and propose an effective fine-tuning procedure tailored for LLMs to generate coherent translations of literary novels.
\end{itemize}


\begin{figure}[!t]
\centering
\includegraphics[width=\textwidth]{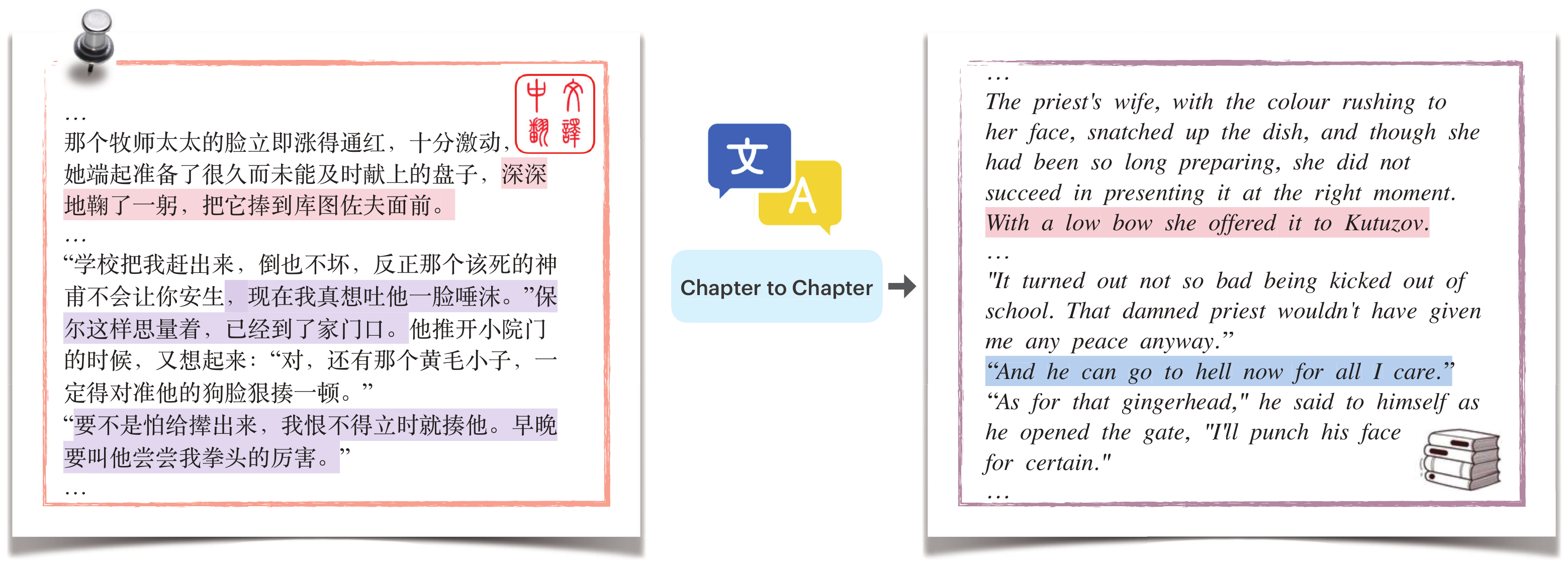}
\caption{An example of of \textsc{Ch2Ch} translation. Sentence Misalignment: \hlc[red!20]{Red} parts are where a source sentence is separated into multiple sentences in the corresponding translation; \hlc[cyan!20]{blue} parts are added by translators and do not have a corresponding source segment; \hlc[violet!20]{violet} parts are deleted by translators in translation.}
\label{fig:unaligned-para}
\vspace{-2mm}
\end{figure}
\section{Preliminary Background}

\subsection{Context-aware Neural Machine Translation}
\paragraph{Sentence-aligned Translation}
In the sentence-aligned setting of context-aware machine translation, we assume that the source and target sentences in a parallel document are well-aligned.
Formally, given a document $D$ comprising a set of source sentences $\bm{X} = \{\bm{x}_1, \bm{x}_2,..., \bm{x}_{d}\}$, there are the same number of sentences $\bm{Y} = \{\bm{y}_1, \bm{y}_2,..., \bm{y}_{d}\}$ in the target side, which are aligned with sentences in $\bm{X}$ by the indices. 
The context-aware neural machine translation (NMT) model computes the probability of translating the source sentence $\bm{x}_i$ conditioned on the context $C_i$, wherein $0 \leq i \leq d$:
\begin{equation}
P_{\textrm{SentAlign}}(\bm{y}_i|\bm{x}_i, \bm{C}_i, \theta) = \prod^{N}_{j=1} P (y_i^{j} | y_i^{<j}, \bm{x}_i, C_{i};\theta).
\vspace{-1mm}
\end{equation}
where $C_{i}$ are contextual sentences surrounding $\bm{x}_i$ and/or  $\bm{y}_i$.
As illustrated in \autoref{fig:unaligned-para}, sentence-aligned translation does not accurately represent real-world translation scenarios.

\paragraph{Paragraph-to-Paragraph Translation}
To get rid of the assumption of sentence-level alignments and leverage richer contextual information, recent work~\citep{thai-etal-2022-exploring,jin-etal-2023-challenges} proposed a paradigm shift towards paragraph-to-paragraph (\textsc{Para2Para}) translation to relax the alignment assumption from sentence-level to paragraph-level.
Concretely, a document $D$ contains a set of aligned parallel paragraphs, $\bm{X} = \{\bm{X}_1, \bm{X}_2,..., \bm{X}_{d}\}$ and $\bm{Y} = \{\bm{Y}_1, \bm{Y}_2,..., \bm{Y}_{d}\}$. Each pair of aligned paragraphs $\bm{X}_i$ and $\bm{Y}_i$ do not necessarily contain the same number of sentences:
\begin{equation}
P_{\textrm{Para2Para}}(\bm{Y}_i|\bm{X}_i, \theta) = \prod^{N}_{j=1} P(Y_i^j | Y_i^{<j}, \bm{X}_i; \theta)
\end{equation}
where $Y_i^{<j}$ are the all previously translated tokens in a paragraph. 
However, in literary texts the splits of paragraphs are equivocal, which limited the application of \textsc{Para2Para} translation to real-world scenario.

\vspace{-1mm}
\subsection{Datasets}
\vspace{-1mm}
Most commonly used corpora, including IWSLT-17~\citep{cettolo-etal-2012-wit3}, NewsCom~\citep{tiedemann-2012-parallel}, Europarl~\citep{koehn-2005-europarl}, and OpenSubtitles~\citep{lison-etal-2018-opensubtitles2018} are sourced from news articles or parliamentary proceedings.
Until recently, some document-level parallel corpora of literary texts have been released. 
\citet{jiang-etal-2023-discourse} curated Bilingual Web Books (BWB), a sentence-aligned corpus that retains document-level information. 
BWB contains 9.6 million sentence pairs sourced from Chinese web novels and their corresponding English translations. 
However, BWB still follows the sentence-level alignment constrains. To support \textsc{Para2Para} translation, 
\citet{thai-etal-2022-exploring} introduced \textsc{Par3}, a paragraph-aligned corpus obtained through both human and automatic translators, containing multilingual non-English novels and their English translations. Another paragraph-aligned corpus, introduced by \citet{al-ghussin-etal-2023-exploring}, consists of parallel paragraphs extracted from Paracrawl~\citep{banon-etal-2020-paracrawl} using automatic sentence alignments. This corpus includes data crawled from the Internet spanning various domains.

\vspace{-1mm}
\subsection{Translation with Large Language Models}
\vspace{-1mm}
LLMs are not explicitly trained on parallel data for translation, yet they possess a profound understanding of languages and can produce coherent text, serving as a valuable foundation for translation tasks~\citep{li2024eliciting}. 
Particularly for resource-rich languages, colossal models with decoder-only architecture, such as GPT-4~\citep{openai2024gpt4}, have approached or even exceeded traditional encoder-decoder models on sentence-level benchmarks and can generate more coherent and human-like translations drawing upon their extensive comprehension of both languages~\citep{robinson-etal-2023-chatgpt, hendy2023good}. 
\citet{xu2023paradigm} proposed a two-stage procedure to finetune Llama2-7b~\citep{touvron2023llama} with a small amount of sentence-level parallel data and obtained impressive improvements over standard sentence-level NMT baselines without LLMs.
\vspace{-1mm}

\section{\textsc{JAM}: Chapter-Aligned Literary Translation Dataset}
\label{sec:data}



\begin{table*}[htbp]
\centering
\resizebox{\textwidth}{!}{
\begin{tabular}{p{4in}p{4in}}
\toprule
\textbf{Source} & \textbf{Target} \\ 
\midrule
“To think what we have been brought to!” Kutuzov cried suddenly, in a voice full of feeling, Prince Andrey's story evidently bringing vividly before him the position of Russia.\newline\newline
“Wait a bit; wait a bit!” he added, with a vindictive look in his face, and apparently unwilling to continue a conversation that stirred him too deeply, he said:\newline\newline
“I sent for you to keep you with me.”   &  
\begin{CJK}{UTF8}{gbsn}
\vspace{3mm}
“弄到什么地步……到什么地步！”库图佐夫突然说，他声音激动，显然，从安德烈公爵的叙述中，他清楚地想象到俄国目前的处境。“给我一段时间，给我一段时间！”他脸上带着愤怒的表情又说，很明显，他不愿继续这个使他激动的话题，他说：“我叫你来，是想让你留在我身边。”
\end{CJK}\\
\midrule
“We must, if everyone wants to; there is no help for it … But, mark my words, my dear boy! The strongest of all warriors are these two—time and patience. They do it all, and our wise counsellors n'entendent pas de cette oreille, voilà le mal. Some say ay, and some say no. What's one to do?” he asked, evidently expecting a reply. “Come, what would you have me do?” he repeated, and his eyes twinkled with a profound, shrewd expression. “I'll tell you what to do,” he said, since Prince Andrey did not answer. “I'll tell you what to do. Dans le doute, mon cher”—he paused—“abstiens-toi.” He articulated deliberately the French saying.   &
\begin{CJK}{UTF8}{gbsn}
“打一仗是可以的，如果大家都愿意的话，没有什么可说的……可是要知道，亲爱的朋友：没有比忍耐和时间这两个战士更强的了，这两位什么都能办成。可是顾问们不肯听这个，困难就在这里。一些人要这样，另一些又不这样。怎么办呢？”他问，显然在等着回答。\newline\newline
“你说说看，我怎么办？”他重复着，眼睛显得深沉、睿智。\newline\newline
“我告诉你怎么办。如果你犹豫不决，亲爱的，”他停了一下，“那你先干别的。”他慢条斯理地一字一句地说。
\end{CJK}\\
\bottomrule
\end{tabular}}
\caption{Examples of paragraph misalignment. Each line represents an individual paragraph in the original text.}
\label{tab:para_misalign_examples}
\vspace{-3mm}
\end{table*}

\vspace{-1mm}

\subsection{Chapter-to-Chapter Translation}
\label{chapt-align}

\vspace{-1mm}

In literary texts, the lengths of paragraphs vary and the splits of paragraphs are equivocal, particularly when dialogues are involved. 
For instance, in novels, dialogue lines are often presented as separate paragraphs, making it challenging to ensure accurate translations without access to the preceding context.
As illustrated by the two examples shown in \autoref{tab:para_misalign_examples}, there are instances where multiple paragraphs from the source side are merged into one paragraph on the target side, and vice versa.

To address this issue, we propose \emph{chapter-to-chapter} (\textsc{Ch2Ch}) translation, a pragmatic and challenging setting, by extending context-aware translation to chapter-level. 
Comparing to paragraph-level alignments, chapter-level alignments provide the model with more comprehensive context from both the source and target texts. This richer context theoretically offers greater potential for improvements and helps mitigate issues such as tense mismatches, particularly in languages like Chinese that lack explicit tense markers~\citep{sun2020rethinking}.

To conduct experiments and facilitate future research endeavours on \textsc{Ch2Ch} translation, we curate a chapter-aligned dataset of English-Chinese literature, named JAM, which comprises 160 English classic novels alongside professional Chinese translations.
In professional literary translation, translators often leverage contexts to enhance the fluency and readability of the translation. To this end, translations may not strictly adhere to sentence alignment\footnote{In 50 sampled paragraphs from JAM there are 18 paragraphs with sentence mis-alignments.}, and some typical sentence misalignment types are listed below, an example is shown in~\autoref{fig:unaligned-para} illustrates:

\begin{description}
    \item \colorbox{cyan!20}{\textcolor{cyan}{\bf{\textsc{Insert}}}}~: new sentence(s) is added by translators and does not have a corresponding source segment.
    
    \item \colorbox{violet!20}{\textcolor{violet}{\bf{\textsc{Delete}}}}~: a source sentence(s) is deleted by translators in translation.
    
    \item \colorbox{red!10}{\textcolor{red}{\bf{\textsc{Split}}}}~: a source sentence is separated into multiple sentences in the corresponding translation.
\end{description}

As such, chapter-to-chapter(\textsc{Ch2Ch}) translation is challenging in nature, given that chapters typically are lengthy and contain complex discourse structure. Detailed experimental results and analysis are provided in Section \ref{ch2ch-challenge}.

\subsection{Data Construction and Quality Control}
\label{data_quality}
\begin{wraptable}{r}{0.5\textwidth}
\vspace{-2mm}
\centering
\resizebox{0.5\textwidth}{!}{%
\begin{tabular}{lccc}
\toprule
  & \multicolumn{1}{l}{\textsc{Chap. \#}} & {\textsc{\begin{tabular}[c]{@{}c@{}}Sentence \#\\ (En/Zh)\end{tabular}}} & \textsc{{\begin{tabular}[c]{@{}c@{}}Word \#  \\ (En/Zh)\end{tabular}}}\\
 \midrule
\textsc{Train} & 4484 & 451.4K / 577.5K & 8.6M / 9.8M\\
\textsc{Valid} & 546 & 52.5K / 68.1K & 1.0M / 1.1M\\
\textsc{Test} & 343 & 44.5K / 55.2K & 814.9K / 955.9K\\
\midrule
\rowcolor{violet!20}
\textsc{Total} & 5373 & 548.5K / 700.9K & 10.4M / 11.9M\\
\bottomrule
\end{tabular}
}
\caption{ JAM Corpus Statistics.}
\label{tab:dataset}
\vspace{-3mm}
\end{wraptable}

We collect 160 bilingual literary books across different genres from the Internet, and format data by manually correcting chapter-level alignment\footnote{We select literary works with chapter breaks, then manually check the alignments of first and last paragraphs for each chapter.}.
Subsequently, we perform standard data cleaning steps (e.g. punctuation normalization) and filter the chapter pairs with a sequence length ratio~$>3.0$. The refined dataset contains a total of 5373 aligned chapters.
The statistics of this dataset are shown in ~\autoref{tab:dataset}~\footnote{English sentences are split by white space; Chinese sentences are segmented using the open-sourced \href{https://github.com/fxsjy/jieba}{Jieba} package.}, and detailed corpus information is in Appendix \ref{app:corpus_statistics}.
The dataset is split into train, valid, and test sets. We randomly select 18 books as the test set. The remaining corpus of 5030 chapters from 142 books was then split into an 80\% training set and a 20\% validation set. 

\section{Experimental Setup}

\subsection{Baselines}
\begin{figure}[!t]
     \centering
     \includegraphics[width=\linewidth]{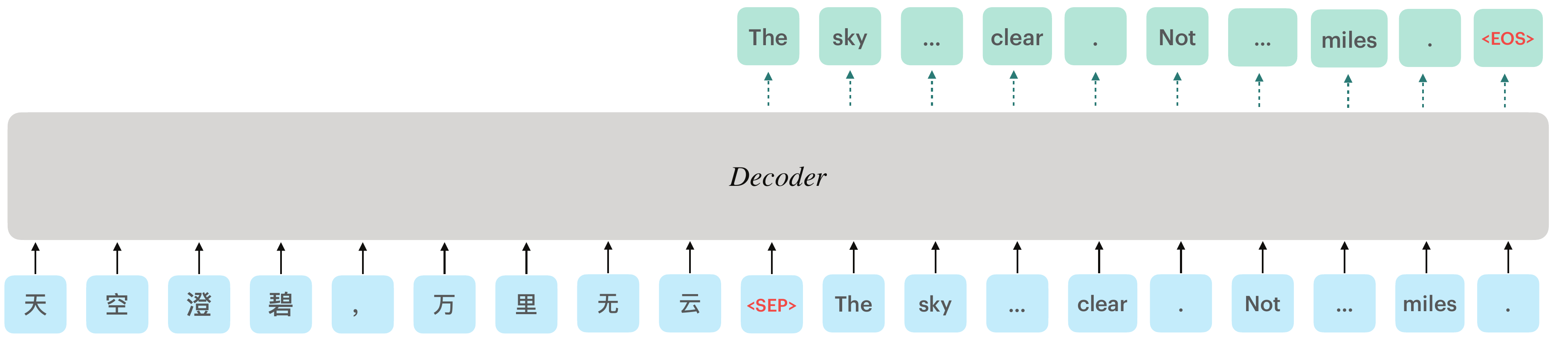}
     \caption{Decoder-only architecture.}
     \label{fig:decoder_example}
\end{figure}

To examine the inherent capacity of the model in the translation task, we perform a benchmarking analysis against two baseline categories:
\vspace{-1mm}
\paragraph{Encoder-Decoder Architecture} 
We use the Transformer~\citep{Vaswani2017} \texttt{base} version, which consists of 6 encoder layers, 6 decoder layers, a model dimension of 512, and an FFN hidden dimension of 2048. 
\vspace{-1mm}
\paragraph{Decoder-only Architecture} 
Compared to the prevalent encoder-decoder architecture, the decoder-only framework is often simpler in architecture and computationally efficient~\citep{fu2023decoderonly}. In the \textsc{Ch2Ch} translation task, we train the decoder-only model using sequences where each source chapter is concatenated with its corresponding target chapter,  demarcated by a \texttt{<SEP>} token, and ended with an \texttt{<EOS>} token:

\begin{center}
    \texttt{<SRC Chapter> <SEP> <TGT Chapter> <EOS>}
\end{center}
The model architecture is shown in \autoref{fig:decoder_example}. 

Motivated by \citet{zhang-etal-2018-improving}, we experiment with training a baseline model on the JAM dataset from scratch, as well as incorporating pre-trained baselines, in which the model is first trained on the sentence-level WMT22 Zh$\xrightarrow{}$En dataset~\citep{kocmi-etal-2022-findings}, before further fine-tuning on the JAM dataset.

\begin{wrapfigure}{r}{0.45\textwidth}
\centering
\vspace{-5mm}
\includegraphics[width=0.45\textwidth]{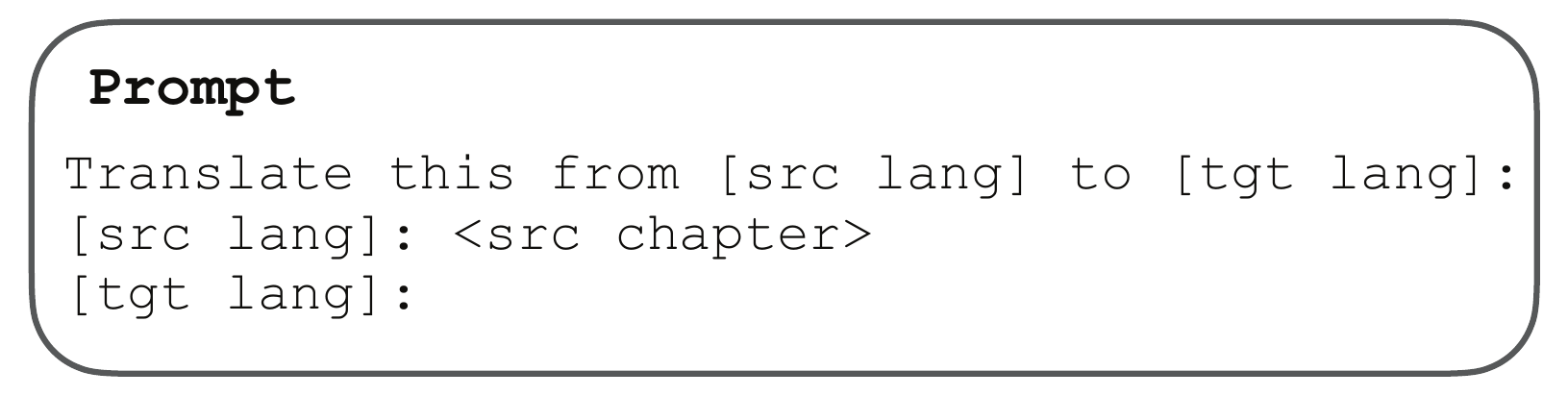}
\caption{Prompt template for LLMs.}
\label{fig:prompt_example}
\vspace{-3mm}
\end{wrapfigure}
\paragraph{Zero-shot Evaluation}
Recent work has showcased the proficiency of LLMs in sentence-level translation. To further probe the ability of LLMs in translating literary, we randomly sample 63 chapters from JAM test set and conduct a zero-shot evaluation on the sampled instances to compare with the following models:
\begin{description}
    \item \colorbox{orange!10}{\textcolor{orange}{\bf{\textsc{NLLB-200-3.3b}}}}~\citep{nllbteam2022language}: an encoder-decoder LLM, with 3.3b parameters. 
    
    \item \colorbox{red!10}{\textcolor{red}{\bf{\textsc{Llama2-7b}}}}~\citep{touvron2023llama}: a generative text model with 7b parameters. 
    
    \item \colorbox{cyan!10}{\textcolor{cyan}{\bf{ALMA-7B}}}~\citep{xu2023paradigm}: finetuned on 5 language pairs from Llama2-7b for translation.
    
    \item \colorbox{violet!10}{\textcolor{violet}{\bf{\textsc{GPT-4}}}}~\citep{openai2024gpt4}: a pre-trained large-scale multi-modal model.
\end{description}
Building upon the approach proposed by~\citet{xu2023paradigm}, we prepend a fixed prompt (see \autoref{fig:prompt_example}) to each chapter.

\paragraph{Finetuning}
We select \textsc{ALMA-7b} to finetune on JAM because of its impressive gains in translation tasks compared to other LLMs; its fine-tuning process is divided into two phrases: first, ALMA-7B-Stage1 finetuned \textsc{Llama2-7b} exclusively on monolingual data; then, the second stage ALMA-7B-Stage2 is subsequently finetuned on parallel data. Specifically, we finetune ALMA-7B-Stage1 on JAM to investigate whether pretraining with sentence-level parallel data is beneficial prior to fine-tuning on chapter-level data.
We use causal language modeling (CLM) loss for finetuning and restrict loss computation only to the target tokens. 

\vspace{-1mm}
\subsection{Handling Long Chapters in Training and Decoding} 
\label{data}
\vspace{-1mm}
As some chapters exceed the maximal context length of some models, we equally segment those chapters into chunks, ensuring that each chunk contains less than 2048 tokens in both Zh and En sides. Data and pre-processing details are in Appendix \ref{app:train-data}.

During decoding, we also pack the maximum number of sentences into blocks within 2048 tokens. The model does not know how many sentences to generate in advance and decoding stops when \texttt{<EOS>} is predicted. 
As illustrated in \autoref{fig:decoder_example}, \texttt{<EOS>} in our experiments is used to indicate the end of translation, not the end of a sentence.

\vspace{-1mm}
\subsection{Post-processing \& Evaluation}
\vspace{-1mm}
Before evaluation, we employ a sliding window with a length of 10 words, calculating the hash value of the substring within the window. As we slide the window, if the hash value of the current substring matches any previously seen hash value, we compare the actual substrings to confirm the repetition and then trim accordingly\footnote{Most repetitions exhibit a self-reinforcement effect, continuously repeating the same sentences or phrases. Therefore, once a repetition is detected, we remove all subsequent words.}. After cleaning, the blocks belonging to the same chapter are merged back together for evaluation at the chapter level.

For all tasks, we report both sentence-level (e.g., BLEU~\citep{papineni-etal-2002-bleu} and COMET~\citep{rei-etal-2020-comet}) and document-level automatic metrics in evaluation. In particular, we analyze the translation quality of LLMs related to specific discourse phenomena such as pronoun ellipsis, named entity coreference by BlonDe score~\citep{jiang2022blonde}.

\vspace{-1mm}
\section{Experimental Result and Analysis}
\vspace{-1mm}
In this section, we report results of our experiments and conduct thorough empirical analysis over a range of model architectures, datasets and decoding strategies. 

\vspace{-1mm}
\subsection{Chapter-to-Chapter Machine Translation Task is Challenging in Nature. }
\label{ch2ch-challenge}
\vspace{-1mm}

Motivated by \citet{zhang-etal-2018-improving}, we experiment with training a baseline model on the JAM dataset from scratch, as well as incorporating a two-stage training procedure, in which the model is first trained on the sentence-level WMT22 Zh$\xrightarrow{}$En dataset~\citep{kocmi-etal-2022-findings}, before further fine-tuning on the JAM dataset.

As illustrates in \autoref{tab:result}, Encoder-Decoder and Decoder-only Transformer models trained from scratch on JAM significantly under-perform the models trained with the 2-stage procedure. 
The significant performance gap demonstrates the challenging nature of \textsc{Ch2Ch} (e.g., 1.87 and 1.09 on BLEU), i.e., the inherent difficulty of training on chapter-level, long-sequence data. 
Translation models that trained with the 2-stage procedure to leverage the sentence-level \texttt{WMT22} exhibit a notable improvement, attesting the difficulty of the \textsc{Ch2Ch} translation task.


\begin{table*}[!t]
\centering
\resizebox{1.0\textwidth}{!}
{%
\begin{tabular}{lccccccccc}
\toprule
\bf{Model} & \bf{WMT22} & \bf{JAM}  & \multicolumn{1}{c}{\bf{BLEU}} & \multicolumn{5}{c}{\bf{BlonDe}} & \bf{COMET}
\\ 
\midrule
& & & & all  & pron. & entity & tense & d.m. & \\ \midrule
Encoder-Decoder & \textcolor{red}{\ding{55}} & \textcolor{green}{\ding{51}} & 1.87 & 8.70 & 49.23 & 19.22 & 42.30 & 17.21 & 0.4128 \\  
Decoder-only & \textcolor{red}{\ding{55}} & \textcolor{green}{\ding{51}}    & 1.09 & 7.23 & 47.46 & 20.77 & 40.40 & 16.54 & 0.4187 \\  
\midrule
Encoder-Decoder & \textcolor{green}{\ding{51}} & \textcolor{green}{\ding{51}} & 14.38 & 31.08 & 
\cellcolor{blue!15}\bf{89.78} & 11.36 & \cellcolor{blue!15}\bf{86.88} & \cellcolor{blue!15}\bf{81.96} & 0.6617\\  
Decoder-only & \textcolor{green}{\ding{51}} & \textcolor{green}{\ding{51}}  & 13.35 & 30.06 & 84.28 & 14.59 & 80.23 & 76.81 & 0.6377\\  

\midrule
ALMA-7B-Stage1 & \textcolor{red}{\ding{55}} & \textcolor{green}{\ding{51}}  & 15.70 & 33.46 & 74.28 & 30.62 & 70.11 & 71.72 & 0.7806 \\

ALMA-7B-Stage2 & \textcolor{red}{\ding{55}} & \textcolor{green}{\ding{51}}  & \cellcolor{blue!15}\bf{16.80} & \cellcolor{blue!15}\bf{35.05} & 78.35 & 
\cellcolor{blue!15}\bf{32.37} & 73.3 & 73.29 & \cellcolor{blue!15}\bf{0.7812} \\
\bottomrule
\end{tabular}}
\caption{Automatic metric results on JAM test set. Note here chapters are segmented by maximum 2048 tokens. ALMA-7B-Stage1 is only fine-tuned on monolingual data. ALMA-7B-Stage2 fine-tunes ALMA-7B-Stage1 on high-quality parallel data. (\textcolor{red}{\ding{55}}) denotes no fine-tuning on corresponding dataset; (\textcolor{green}{\ding{51}}) denotes fine-tuning. \hlc[blue!15]{\textbf{Bold}} denotes best performance.}
\label{tab:result}
\end{table*}

\begin{figure}[t]
\centering
\includegraphics[width=\textwidth]{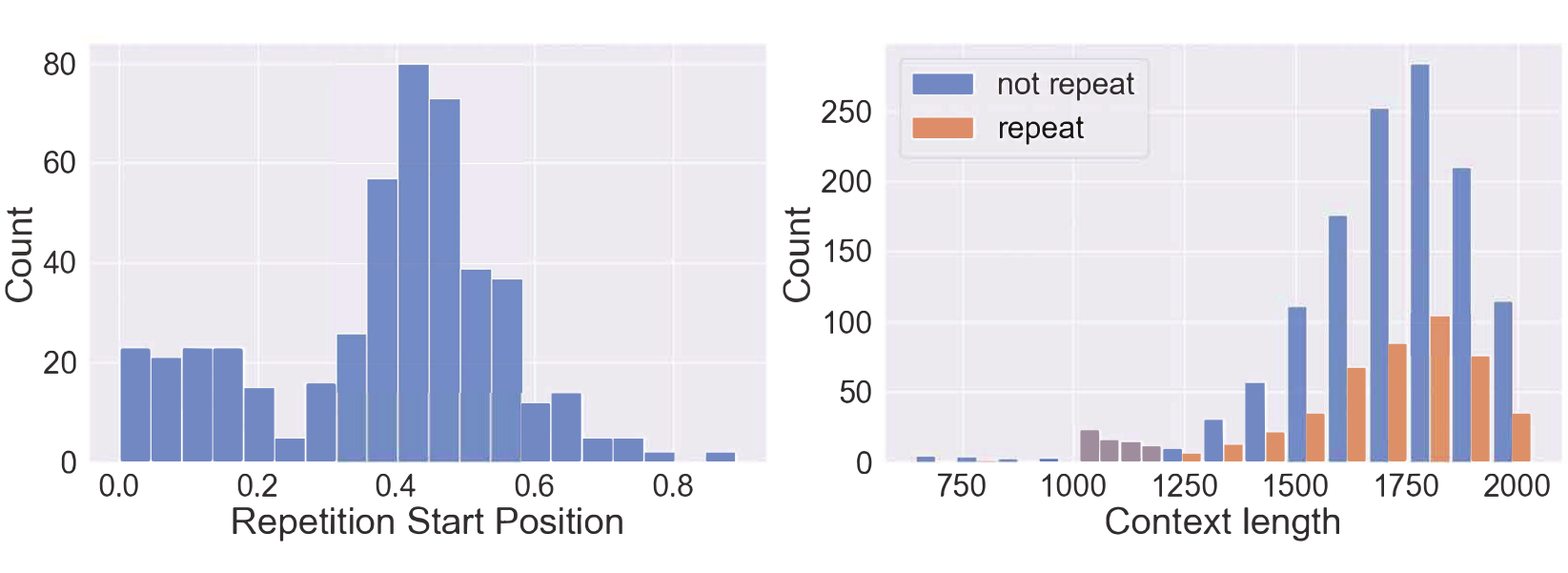}
\caption{Left: Repetition start position in each sentence; Right: Repetition distribution across various context length}
\label{fig:repetition_analysis}
\vspace{-3mm}
\end{figure}

\subsection{Effective Fine-tuning and Decoding Strategy}
\label{finetuning}

\paragraph{Does sentence-level fine-tuning help?}
We next investigate the prerequisite of sentence-level fine-tuning prior to the training on JAM dataset by comparing ALMA-7B-Stage1 and ALMA-7B-Stage2 respectively, with the latter has been fine-tuned on sentence-level parallel datasets. 
\autoref{tab:result} indicates that such sentence-level fine-tuning improves BLEU from 15.7 to 16.8 and BlonDe from 33.46 to 35.05, suggesting that fine-tuning at sentence-level contributes positively to the accuracy of literary translation. 
In contrast, the improvement on COMET is marginal, possibly attributable to COMET's focus on assessing the coherence and fluency of the generated translations. 
These qualities might already be sufficiently robust in an LLM.

\paragraph{Repetition Problem in Translation Decoding}
\citet{deutsch2023training} founds that translation does not degrade as the sequence becomes longer. 
However, according to our results, this is not universally the case; the effectiveness of translation diminishes as the context becomes really lengthy. 
To investigate the insights, we examine the translations of JAM test set on the fine-tuned ALMA-7B-Stage2 model and observe a notable pattern of undesirable repetitions---either phrases or entire sentences---emerges within the generated translations. 

\begin{wrapfigure}{r}{0.45\textwidth}
\centering
\vspace{-1mm}
\includegraphics[width=0.45\textwidth]{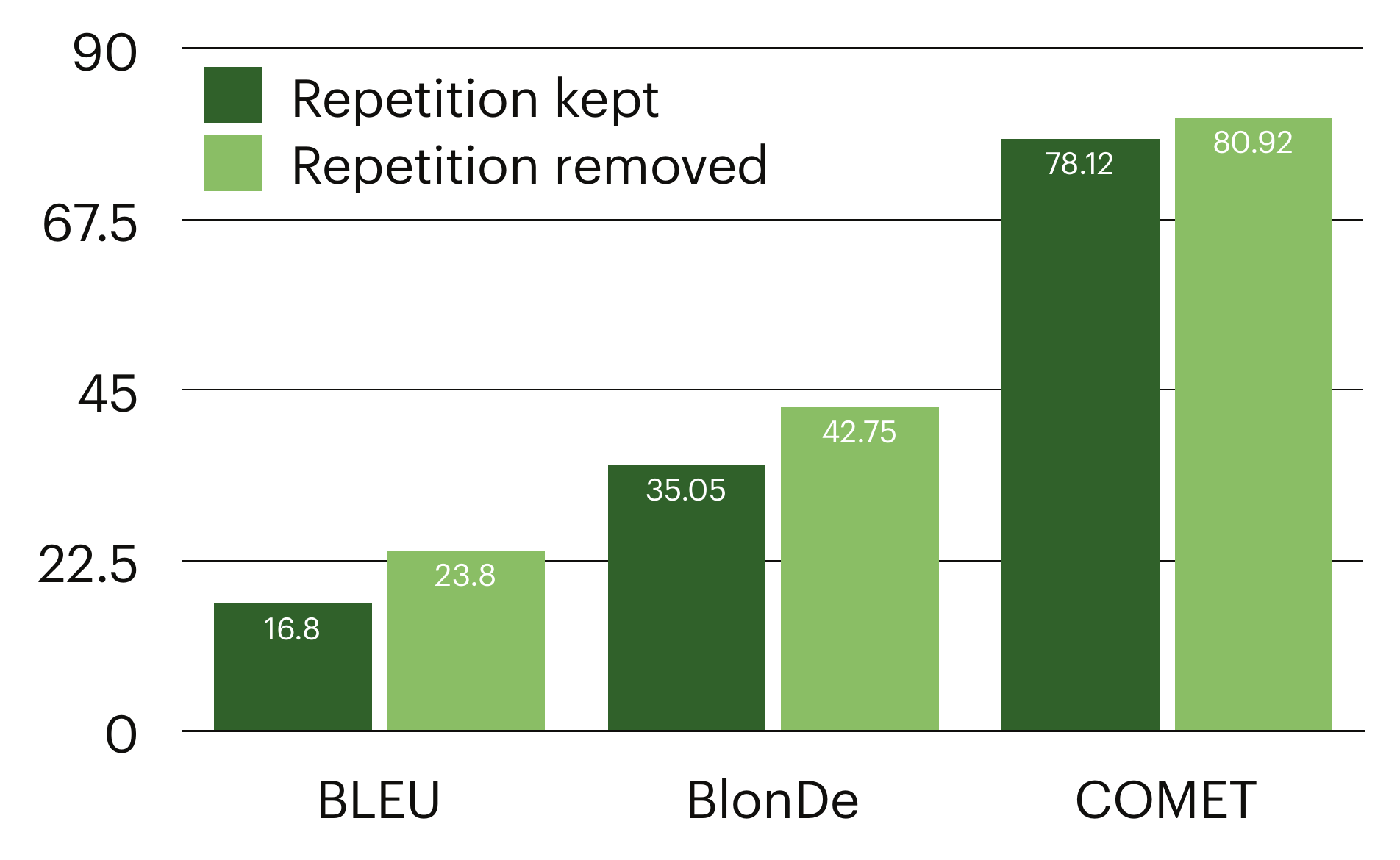}
\captionof{figure}{
Comparison between finetuned \textsc{ALMA-7B} on JAM, with versus w/o post-repetition removal processing.}
\vspace{-7mm}
\label{fig:post-processing}
\end{wrapfigure}

Specifically, 36.7\% of the translations within our test set exhibit some form of repetition. As illustrates in~\autoref{fig:repetition_analysis}, repetition occurs predominantly located within the first half of the translations\footnote{Detailed Blonde scores across different categories are presented in Appendix \ref{app:finetune_repetition}}. Furthermore, sentences exceeding 1000 tokens are more likely to generate repetitive words, phrases or sentences\footnote{We also conduct repetition analysis for all zero-shot generations across various architectures in Appendix \ref{app:zero_shot_repetition}}. 
This observation is consistent with earlier studies indicating text generation with LLMs often results in consecutive sentence-level repetitions, attributed to the use of maximization-based decoding algorithms.\citep{holtzman2020curiouscaseneuraltext,xu2023lookbackdecodingopenendedtext}. The detailed analysis by \citet{xu2022learningbreakloopanalyzing} sheds light on the underlying causes: these models have an inherent tendency to repeat previous sentences, and they tend to overestimate the probability of repeated sequences. This repetition problem is particularly evident in long-context translation, where increasing the chunk length amplifies the risk of the model falling into repetitive loops.

To further evaluate the model's translation ability, we implement post-processing to eliminate repetitions in the generations. 
According to \autoref{fig:post-processing}, this approach enhances translation quality significantly across all metrics.
This leads to a potential direction of future work to develop advanced decoding algorithms to avoid repetitions in translation.





\begin{wraptable}{r}{0.5\textwidth}
\centering
\vspace{-3mm}
\resizebox{0.5\textwidth}{!}{%
\begin{tabular}{clccc}
\toprule
\textbf{Ft.} & \textbf{Decoding} & \textbf{BLEU} & \textbf{BlonDe} & \textbf{COMET}  \\
\midrule
 \textcolor{red}{\ding{55}} & Greedy &  3.7 & 11.81 & 0.6012 \\
 \textcolor{red}{\ding{55}} & Beam-5 &  2.7 & 9.09 & 0.5433   \\
 \textcolor{green}{\ding{51}} & Greedy &  14.0 & 31.26 & 0.7806  \\
 \textcolor{green}{\ding{51}} & Beam-5 &  \cellcolor{violet!20}\bf{16.8} & \cellcolor{violet!20}\bf{35.05} & \cellcolor{violet!20}\bf{0.7812}   \\
\textcolor{green}{\ding{51}} &  + \textit{rp} &  \bf{19.1} & \bf{37.25} & \bf{0.8028} \\
\bottomrule
\end{tabular}
}
\caption{Comparison of decoding strategies across different evaluation metrics of \textsc{ALMA-7B} performance. (\textcolor{red}{\ding{55}}) No fine-tuning on \textsc{JAM} dataset; (\textcolor{green}{\ding{51}}) denotes fine-tuning. \textit{rp} denotes repetition penalty=1.18 }
\label{tab:decoding}
\vspace{-2mm}
\end{wraptable}
\vspace{-2mm}
\paragraph{Comparison of Decoding Strategies}
\label{analysis:decoding}
By default, beam search is employed for all models, with beam size 5. However, upon training certain LLMs on the \textsc{Ch2Ch} task, we observe sub-optimal performance with beam search.  We investigate the performance of two decoding strategy: \textit{greedy decoding} and \textit{beam search} decoding through a fine-grained analysis on the JAM test set.
\autoref{tab:decoding} presents the experimental results. Greedy decoding poses as a weak methodology and its presence has not been found to substantially boost translation performance compared with Beam search.

\vspace{-1mm}
\subsection{How Do Large Language Models Perform on Literary Translation?}
\label{analysis:llms}
\vspace{-1mm}

\begin{figure}[!t]
\centering
\includegraphics[width=1.0\textwidth]{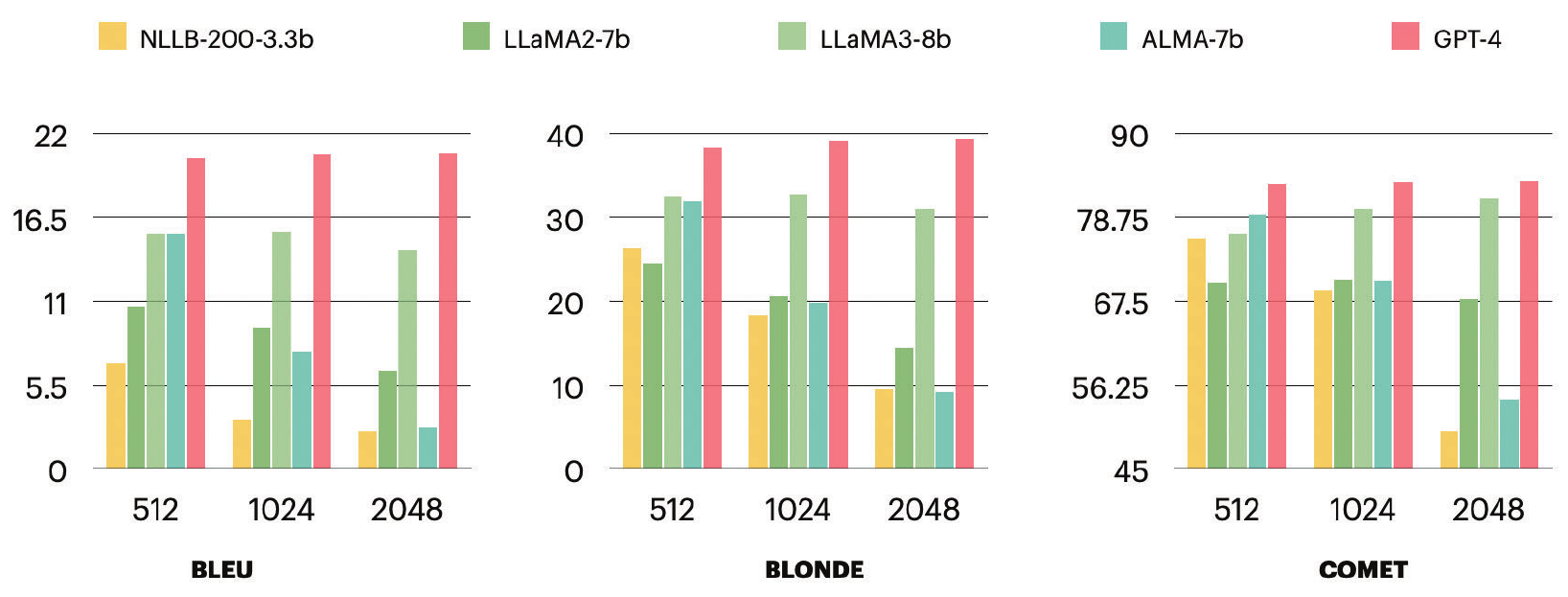}
\caption{Zero-shot performance on JAM data across LLMs. The chapter-level data are
segmented into chunks containing at most 512, 1024, 2048 tokens. ACL = average chapter
length in tokens; The ACL of sampled instances=1850.}
\label{fig:zero-shot-chart}
\vspace{-3mm}
\end{figure}

In order to evaluate the capacity of LLMs on \textsc{Ch2Ch} translation , we perform zero-shot evaluation on the JAM dataset across different models. To further analyze performance variations across different context lengths, we segment chapters into at most 512, 1024, and 2048 tokens, respectively. 
The results are presented in \autoref{fig:zero-shot-chart}. 

\textsc{GPT-4} outperforms all other models across both sentence-level and document-level metrics. 
Rather, translation models with less parameters, such as \textsc{NLLB-3.3b} and ALMA-7B-Stage2, struggle in the \textsc{Ch2Ch} task, i.e., performance drop dramatically especially when the sequence become longer than 1024 tokens. 
One reason as to why ALMA-7B-Stage2 faces challenges in translating long sentences is that it has been finetuned exclusively on short parallel sequences.
This may impair its capability to handle long-sequence translation and fully exploit the advantages of chapter-level translation. However, we observe notable improvements after fine-tuning ALMA-7B on our chapter-level dataset JAM even in the most challenging setting where the context extends up to 2048 tokens, as shown in \autoref{tab:result}.


Despite LLMs such as \textsc{Llama2} being theoretically capable of handling contexts of up to 4096 tokens, their performance in translation tasks over extensive contexts remains subpar. Before delving into more nuanced improvements in discourse-level translation, it is crucial to enhance the model's capacity for high-quality long-context translation.


\begin{wrapfigure}{r}{0.45\textwidth}
\centering
\vspace{-5mm}
\includegraphics[width=0.45\textwidth]{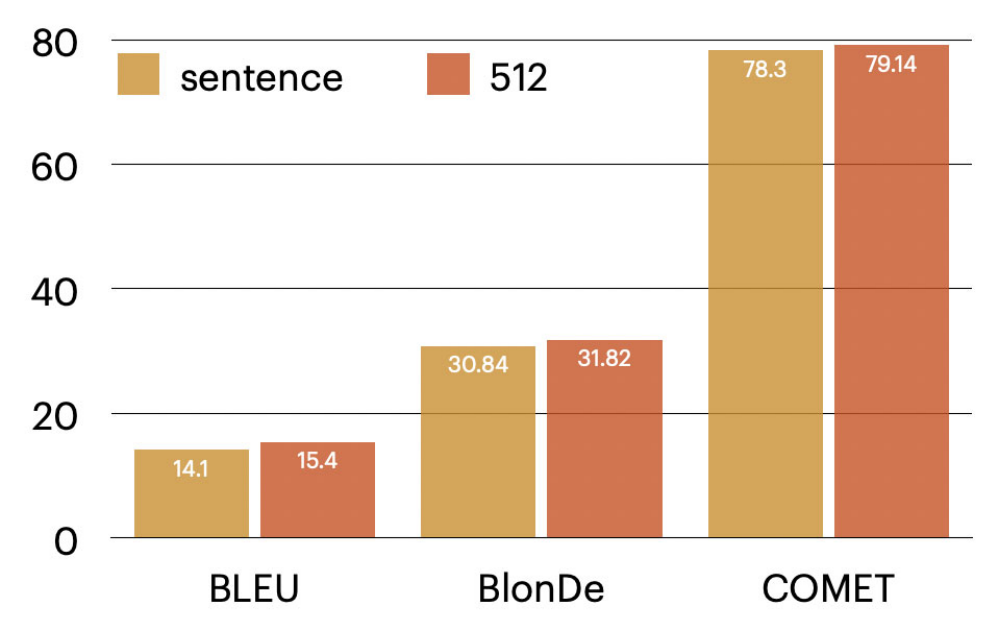}
\caption{Comparison between sentence and 512 tokens segmentation.}
\label{fig:length_comparison}
\vspace{-5mm}
\end{wrapfigure}
\vspace{-2mm}
\paragraph{\textsc{Ch2Ch} vs. Sentence Translation}
The high-level objective of \textsc{Ch2Ch} translation is to leverage more training signals from chapter-level dataset. To test the effectiveness of this setting, we conduct an experiment to segment chapters into sentences for comparison. 
Concretely, we first split each chapter into separated sentences using the NLTK~\footnote{\url{https://github.com/nltk/nltk}} package, then execute translation individually on each sentence with ALMA-7B.
The translated sentences are concatenated back to calculate document-level evaluation metrics. \autoref{fig:length_comparison} indicates that ALMA-7B under the 512-tokens setting outperforms the sentence-segmented setting across all metrics, attesting the significance of \textsc{Ch2Ch} translation.

\paragraph{Decoder-only vs. Encoder-Decoder Architecture}
\label{analysis:architecture}


Under the zero-shot setting (\autoref{fig:zero-shot-chart}), ALMA-7B-Stage2 continues to surpass encoder-decoder translation model \textsc{NLLB-200-3.3B} on BLEU scores. 
In terms of document-level evaluation metrics, ALMA-7B-Stage2 performs on par
with, or even better than \textsc{NLLB-200-3.3B} on the most BlonDe metrics, e.g., pronnoun and discourse marker(d.m.). 
One potential explanation is that the backbone LLM \textsc{Llama2-7b} has a better context understanding and text generating ability. 
For example, discourse markers, e.g., \textit{however, on the other hand}, are crucial for maintaining the coherence and cohesion of text, areas in which LLMs are trained. 
Furthermore, \textsc{NLLB-200-3.3B} tends to generate shorter text compared to other models. 
One hypothesis is that it is primarily trained on a sentence-aligned dataset, where the source and target sentences do not differ significantly in length.

After finetuning on JAM, though Encoder-Decoder perform slightly better than Decoder-only model, yet still under-perform \texttt{ALMA} models on most of the evaluation metrics (\autoref{tab:result}). The above results demonstrates the effectiveness of decoder-only models in handling complex literary translation. Particularly noteworthy is the fact that LLMs do not rely heavily on large amounts of parallel data and are inherently capable of translating long context sequences after finetuning.



\section{Conclusion}
While machine translation demonstrates strong sentence-level performance, it still falls short of human translation in effectively utilizing long-context information. In our paper, we show that Chapter-to-Chapter (\textsc{Ch2Ch}) translation is a viable approach for \textit{context-aware} NMT, exemplified by our novel dataset, JAM. Chapter-level data, derived from professional translations, offers richer context signals and presents a more realistic scenario. Through detailed empirical experiments, we discover that LLMs are aptly suited for \textsc{Ch2Ch} translation following a two-step fine-tuning process: first at the sentence level, then at the chapter level. This procedure equips LLMs with a robust understanding of context, resulting in translations that are both coherent and context-aware. Nevertheless, challenges arise at the chapter level, notably the issue of repetition inheriting from LLMs' long-context generation, signaling the need for improved decoding strategies in future research.





\bibliography{colm2024_conference}
\bibliographystyle{colm2024_conference}

\newpage

\appendix
\section*{Appendix: Towards Chapter-to-Chapter Context-Aware Literary Translation via Large Language Models}

\section{JAM Dataset}

\subsection{Corpus Information}
\label{app:corpus_statistics}
\begin{table*}[ht]
\centering
\resizebox{1.0\textwidth}{!}
{%
\begin{tabular}{llccc}
\toprule
\textbf{Title} & \textbf{Author} & \multicolumn{1}{l}{\textbf{Year}} & \multicolumn{1}{l}{\textbf{\#Chapts}} & \multicolumn{1}{l}{\textbf{ACL (en/zh)}} \\
\midrule
1984                             & George Orwell            & 1949 & 24  & 5.8K/10.2K  \\
A Tale of Two Cities             & Charles Dickens          & 1859 & 44  & 4.3K/8.0K   \\
Ancient Greek Myths              & /                        & /    & 58  & 488.2/862.1 \\
Don Quixote                      & Miguel de Cervantes      & 1605 & 125 & 4.4K/6.9K   \\
How The Steel Was Tempered       & Nikolai Ostrovsky        & 1934 & 18  & 11.7K/24.8K \\
Little Prince                    & Antoine de Saint-Exupéry & 1943 & 28  & 822.3/1.4K  \\
Little Women                     & Louisa May Alcott        & 1868 & 47  & 5.8K/10.7K  \\
Lord of the Flies                & William Golding          & 1954 & 12  & 7.8K/16.8K  \\
Oliver Twist                     & Charles Dickens          & 1838 & 53  & 4.4K/8.7K   \\
Robinson Crusoe                  & Daniel Defoe             & 1719 & 8   & 20.9K/35.4K \\
The Adventures of Tom Sawyer     & Mark Twain               & 1876 & 35  & 3.1K/5.7K   \\
The Giver                        & Lois Lowry               & 1993 & 23  & 2.8K/5.3K   \\
The Shawshank Redemption         & Stephen King             & 1982 & 35  & 1.6K/2.7K   \\
Wuthering Heights                & Emily Brontë             & 1847 & 34  & 5.1K/9.3K   \\
The Time Machine                 & H. G. Wells              & 1895 & 13  & 3.4K/6.2K   \\
Alice’s Adventures in Wonderland & Lewis Carroll            & 1865 & 9   & 3.1K/5.7K   \\
The Mysterious Island            & Jules Verne              & 1875 & 62  & 4.5K/8.2K   \\
The Old Man and the Sea          & Ernest Hemingway         & 1952 & 6   & 5.0K/10.3K  \\
Sophies World                    & Jostein Gaarder          & 1991 & 35  & 6.8K/12.6K  \\
Black Beauty                     & Anna Sewell              & 1877 & 13  & 1.9K/3.0K \\ 
\bottomrule
\end{tabular}
}
\caption{Corpus information for 20 sample books. ACL = average chapter length in tokens.}
\label{tab:corpus-info}
\end{table*}

\autoref{tab:corpus-info} shows 20 sample books from the JAM dataset, in which the \textsc{ACL} column is obtained by using \href{https://huggingface.co/docs/transformers/main/model_doc/llama#transformers.LlamaTokenizerFast}{LlamaTokenizerFast}.

\section{Implementation Details}
\label{app:impl}

\subsection{Data}
\label{app:train-data}

Data for baseline models is encoded and vectorized with byte-pair encoding \cite{sennrich-etal-2016-neural} using the \texttt{SentencePiece}~\citep{kudo2018sentencepiece} framework. We use a 32K joint vocabulary size for Zh$\rightarrow$En. Full corpus statistics of WMT22 are in Table \ref{tab:corpora}.

\begin{table}[ht]
\centering
\resizebox{0.5\columnwidth}{!}
{%
\begin{tabular}{lllll}\toprule
Dataset  & Lg. Pair             & Train     & Valid   & Test  \\ \toprule
WMT22   & Zh$\rightarrow$En & 25134743 & 2002 & 2001 \\
\bottomrule
\end{tabular}}
\caption{Sentence counts across WMT22 datasets.}
\label{tab:corpora}
\end{table}

To segment JAM chapter-level dataset into chunks, we first decide the number of chunks to split in a chapter by ensuring that each chunk includes no more than 2048 English and Chinese tokens, then equally segment the chapter into the computed number of chunks. There is no overlap between chunks, and we keep a sentence  a complete unit when we split chapters. 

\subsection{Baseline Traning}
\label{app:baseline-train}

We train baseline models (Encoder-decoder and Decoder-only) on the \texttt{fairseq} framework . Following \citet{Vaswani2017, fernandes-etal-2021-measuring}, we use the Adam optimizer with $\beta_1 = 0.9$ and $\beta_2 = 0.98$, dropout set to 0.3, an inverse square root learning rate scheduler with an initial value of $10^{-4}$, and the warm-up step set to 4000. Here, we only train the Transformer \texttt{base} version, and the decoder-only model is also derived from the base Transformer \texttt{base} architecture. We keep the parameter size of both Encoder-decoder and Decoder-only architecture similar for fair comparison.

\subsection{LLM Training}
All models are trained with 8xA40 GPUs and DeepSpeed+ZeRO3. Following \citet{xu2023paradigm}, we use Adam optimizer, weight decay set to 0.01, and the warm-upratio set to 0.01, an inverse square root learning rate scheduler with an initial value of $2\times 10^{-5}$. 
\begin{table*}[!t]
\centering
\resizebox{\textwidth}{!}
{%
\begin{tabular}{lcccccccc}
\toprule
\bf{Model}  & \multicolumn{1}{c}{\bf{BLEU}} & \multicolumn{5}{c}{\bf{BlonDe}}  & \multicolumn{1}{c}{\bf{COMET}} & \bf{ACL} \\ \midrule
& &  all & pron. & entity & tense & d.m. &  &\\
\midrule
&&&&\textit{512 tokens}&&&& \\ \noalign{\vskip 0.5ex}
NLLB-200-3.3b & 6.90 & 26.37 & 63.26 & 23.96 & 63.53 & 61.59 & 0.7592 & 870\\
LLaMA2-7b & 10.60 & 24.49 & 73.89 & 17.51 & 72.70 & 66.85 & 0.6990 & 1551 \\
ALMA-7b  & 15.40 & 31.82 & 88.35 & 19.69 & 88.22 & 82.30 & 0.7914 & 1608 \\
GPT-4  & \cellcolor{yellow!40}\bf{20.40} & \cellcolor{yellow!40}\bf{38.24} & \cellcolor{yellow!40}\bf{91.03} & \cellcolor{yellow!40}\bf{39.43} & \cellcolor{yellow!40}\bf{90.34} & \cellcolor{yellow!40}\bf{82.35} & \cellcolor{yellow!40}\bf{0.8324} & \cellcolor{yellow!40}\bf{1863} \\
\hdashline\noalign{\vskip 1ex}
&&&&\textit{1024 tokens}&&&&\\ 
NLLB-200-3.3b  & 3.20 & 18.32 & 47.37 & 17.17 & 46.15 & 44.29 & 0.6888 & 709\\
LLaMA2-7b  & 9.30 & 20.57 & 64.09 & 11.60 & 66.44 & 59.74 & 0.7025 & 1648\\
ALMA-7b  & 7.70 & 19.82 & 68.49 & 13.30 & 71.00 & 62.49 & 0.7017  & 2223 \\
GPT-4 & 
\cellcolor{blue!10}\bf{20.60} & 
\cellcolor{blue!10}\bf{39.20} & 
\cellcolor{blue!10}\bf{91.12} & 
\cellcolor{blue!10}\bf{40.87} & 
\cellcolor{blue!10}\bf{90.32} & 
\cellcolor{blue!10}\bf{82.87} & 
\cellcolor{blue!10}\bf{0.8347} & 
\cellcolor{blue!10}\bf{1821} \\
\hdashline\noalign{\vskip 1ex}
&&&&\textit{2048 tokens}&&&& \\ \noalign{\vskip 0.5ex}
NLLB-200-3.3b  & 2.50 & 9.48 & 41.62 & 7.37 & 50.66 & 25.98 & 0.5009 & 1254 \\
LLaMA2-7b  & 6.40 & 14.40 & 49.45 & 8.63 & 53.66 & 39.69 & 0.6778 & 1780\\
ALMA-7b  & 2.70 & 9.09 & 42.27 & 6.35 & 47.98 & 27.77 & 0.5433 & 2382\\
GPT-4 & \cellcolor{red!20}\bf{20.70} & \cellcolor{red!20}\bf{39.35} & \cellcolor{red!20}\bf{91.39} & \cellcolor{red!20}\bf{41.81} & \cellcolor{red!20}\bf{91.39} & \cellcolor{red!20}\bf{83.67} & \cellcolor{red!20}\bf{0.8359} & \cellcolor{red!20}\bf{1765}\\
\bottomrule
\end{tabular}}
\caption{Zero-shot performance on JAM data across LLMs. The chapter-level data are segmented into chunks containing at most 512, 1024, 2048 tokens. ACL = average chapter length in tokens; The ACL of sampled instances=1850. }
\label{tab:zero-shot-result}
\vspace{-3mm}
\end{table*}

The zero-shot evaluation on JAM dataset across different chunk sizes are shown in \autoref{tab:zero-shot-result}.
\begin{figure}[ht]
\centering
\includegraphics[width=0.5\textwidth]{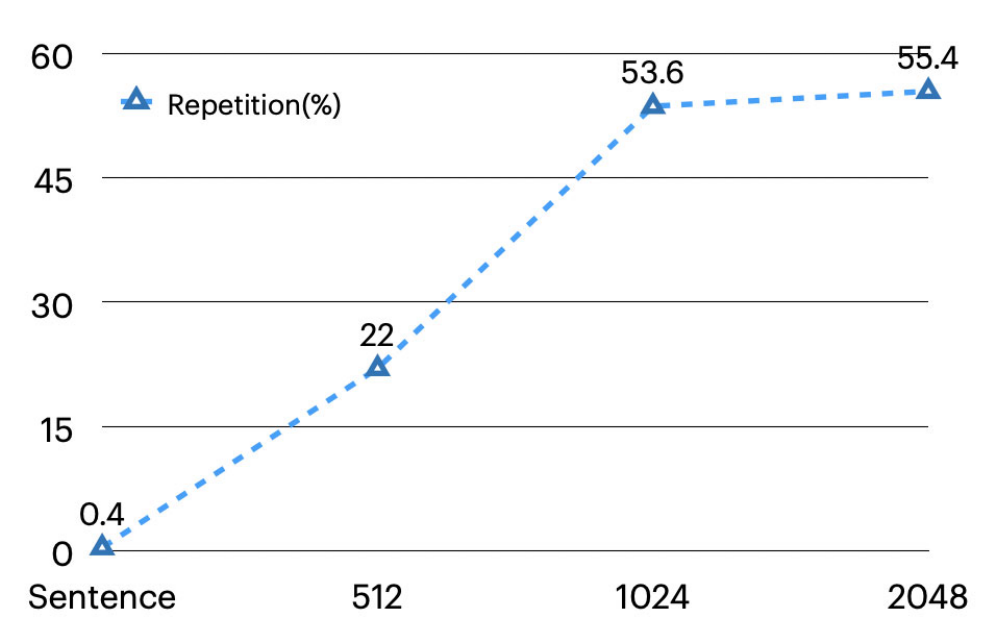}
\caption{Repetition ratio in the generation results for different input context length}
\label{fig:zero-shot-repeat-ratio}
\end{figure}

\subsection{Repetition Analysis on Zero-shot Translations}
\label{app:zero_shot_repetition}
\begin{figure}[ht]
\centering
\includegraphics[width=1.0\textwidth]{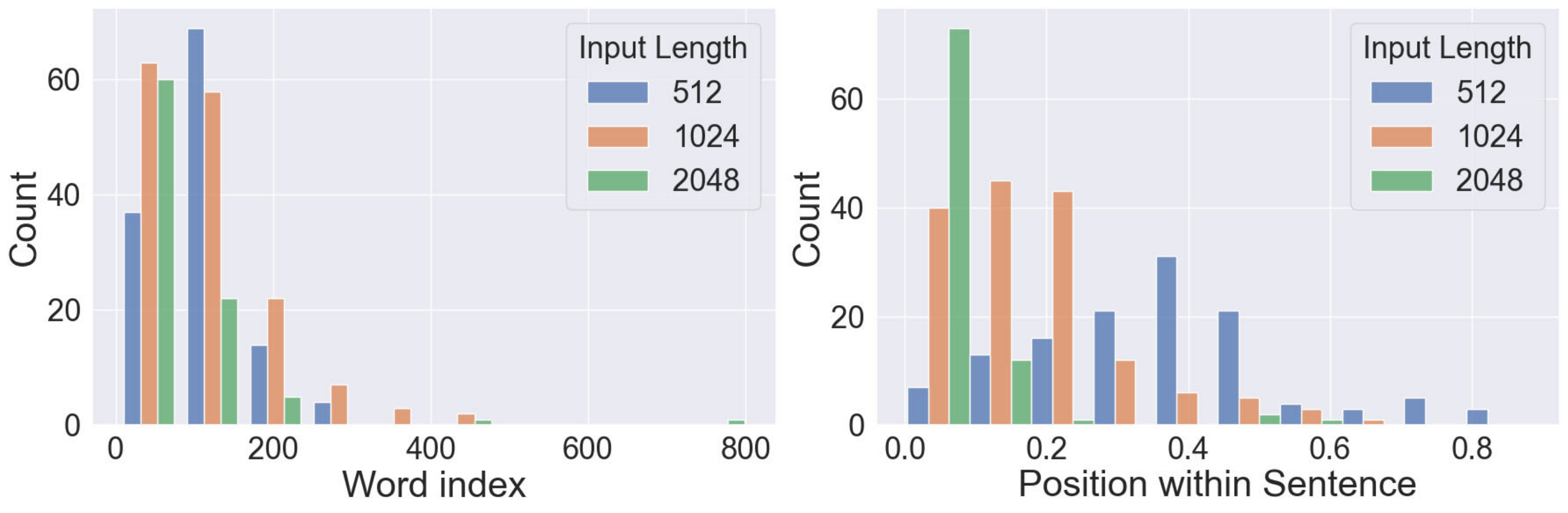}
\caption{Repetition start position across different input lengths. Left: The word index of repetition, Right: The relative position of repetition.}
\label{fig:zero-shot-repeat}
\end{figure}

As illustrated in \autoref{fig:zero-shot-repeat-ratio}, repetition is not an issue for sentence-level translation. However, the repetition ratio significantly increases as the input context length increases from 512 to 1024. Furthermore, \autoref{fig:zero-shot-repeat} shows that as the input length increases, the repetition start position also occurs earlier.

\subsection{Post-processing on Fine-tune Translations}
Post-processing eliminate repeated words and phrases in generated translations. \autoref{tab:repetition-compare} shows a comprehensive automatic metric comparison between translations with post-processing versus. without post-processing. 
\label{app:finetune_repetition}
\begin{table*}[!h]
\centering
\resizebox{1.0\textwidth}{!}
{%
\begin{tabular}{lcccccccccc}
\toprule
\bf{Model} & \bf{WMT22} & \bf{JAM} & \bf{Post-processing} & \multicolumn{1}{c}{\bf{BLEU}} & \multicolumn{5}{c}{\bf{BlonDe}} & \bf{COMET}
\\ 
\midrule
& & &  & & all  & pron. & entity & tense & d.m. & \\ \midrule
ALMA-7B-Stage1 & \textcolor{red}{\ding{55}} & \textcolor{green}{\ding{51}}  & \textcolor{red}{\ding{55}} &  15.70 & 33.46 & 74.28 & 30.62 & 70.11 & 71.72 & 0.7806 \\

ALMA-7B-Stage2 & \textcolor{red}{\ding{55}} & \textcolor{green}{\ding{51}}  & \textcolor{red}{\ding{55}} &  16.80 & 35.05 & 78.35 & 
32.37 & 73.3 & 73.29 & 0.7812 \\
ALMA-7B-Stage1 & \textcolor{red}{\ding{55}} & \textcolor{green}{\ding{51}} & \textcolor{green}{\ding{51}} & 21.6 & 39.54 & 86.43 & 35.43 & 84.52 & 82.98 & 0.7986 \\
ALMA-7B-Stage2 & \textcolor{red}{\ding{55}} & \textcolor{green}{\ding{51}}  & \textcolor{green}{\ding{51}} &
\cellcolor{blue!15}\bf{23.8} & \cellcolor{blue!15}\bf{42.75} & \cellcolor{blue!15}\bf{90.75} & \cellcolor{blue!15}\bf{39.83} & \cellcolor{blue!15}\bf{88.86} & \cellcolor{blue!15}\bf{85.16} & \cellcolor{blue!15}\bf{0.8092}\\
\bottomrule
\end{tabular}}
\caption{Automatic metric result of ALMA-7B translations on JAM, with versus without post repetition removal processing. \hlc[blue!15]{\textbf{Bold}} denotes best performance.}
\label{tab:repetition-compare}
\vspace{-3mm}
\end{table*}

\end{document}